\documentclass{article}

    \usepackage[final, nonatbib]{neurips_2023}

\usepackage[numbers]{natbib}

\usepackage[utf8]{inputenc} 
\usepackage[T1]{fontenc}    
\usepackage{hyperref}       
\usepackage{url}            
\usepackage{booktabs}       
\usepackage{amsfonts}       
\usepackage{nicefrac}       
\usepackage{microtype}      
\usepackage[svgnames]{xcolor}
\usepackage{algorithm}
\usepackage{algpseudocode}
\usepackage{wrapfig}
\usepackage{graphicx}
\usepackage{subcaption}
\usepackage{multicol}

\usepackage{listings}
\lstalias{plantuml}{}

\title{LLM Augmented Hierarchical Agents}

\author{%
  Bharat Prakash\textsuperscript{\rm 1}, Tim Oates\textsuperscript{\rm 1}, Tinoosh Mohsenin\textsuperscript{\rm 1, 2}\\
  University of Maryland, Baltimore County\textsuperscript{\rm 1}\\
  Johns Hopkins University \textsuperscript{\rm 2}\\
}

\begin{document}

\maketitle

\begin{abstract}
Solving long-horizon, temporally-extended tasks using Reinforcement Learning (RL) is challenging, compounded by the common practice of learning without prior knowledge (or \emph{tabula rasa} learning). Humans can generate and execute plans with temporally-extended actions and quickly learn to perform new tasks because we almost never solve problems from scratch. We want autonomous agents to have this same ability. Recently, LLMs have been shown to encode a tremendous amount of knowledge about the world and to perform impressive in-context learning and reasoning. However, using LLMs to solve real world problems is hard because they are not grounded in the current task. In this paper we exploit the planning capabilities of LLMs while using RL to provide learning from the environment, resulting in a hierarchical agent that uses LLMs to solve long-horizon tasks. Instead of completely relying on LLMs, they guide a high-level policy, making learning significantly more sample efficient. This approach is evaluated in simulation environments such as MiniGrid, SkillHack, and Crafter, and on a real robot arm in block manipulation tasks. We show that agents trained using our approach outperform other baselines methods and, once trained, don't need access to LLMs during deployment.
\end{abstract}

\section{Introduction}

Humans can generate and execute plans with temporally extended actions to perform complex tasks in a dynamic and uncertain world.  We would like autonomous agents to have the same capabilities.  Massive engineering efforts can lead to agents that are remarkably robust, such as rovers in space, and surgical and industrial robots.  In the absence of such resources, techniques such as Reinforcement Learning (RL) can be used to extract robust control policies from experience.  
However, RL has many challenges, such as exploration under sparse rewards, generalization, safety, etc. This makes it difficult to learn good policies in a sample efficient way. Popular ways to tackle these problems include using expert feedback \cite{christiano2017deep, Warnell2018} and leveraging the hierarchical structure of complex tasks. There is significant prior work on learning hierarchical policies to break down tasks into smaller sub-tasks \cite{sutton1999between, fruit2017exploration, bacon2017option}. 

Hierarchical Reinforcement Learning (HRL) does indeed mitigate some of the problems mentioned above. However, as the number of options or skills increases, we face some of the same problems again. Using some form of supervision, such as providing details about the sub-tasks or intermediate rewards or high-level human guidance, is one approach \cite{prakash2021interactive, jiang2019language, le2018hierarchical}. 

\begin{figure}
  \centering
  \includegraphics[width=0.9\columnwidth]{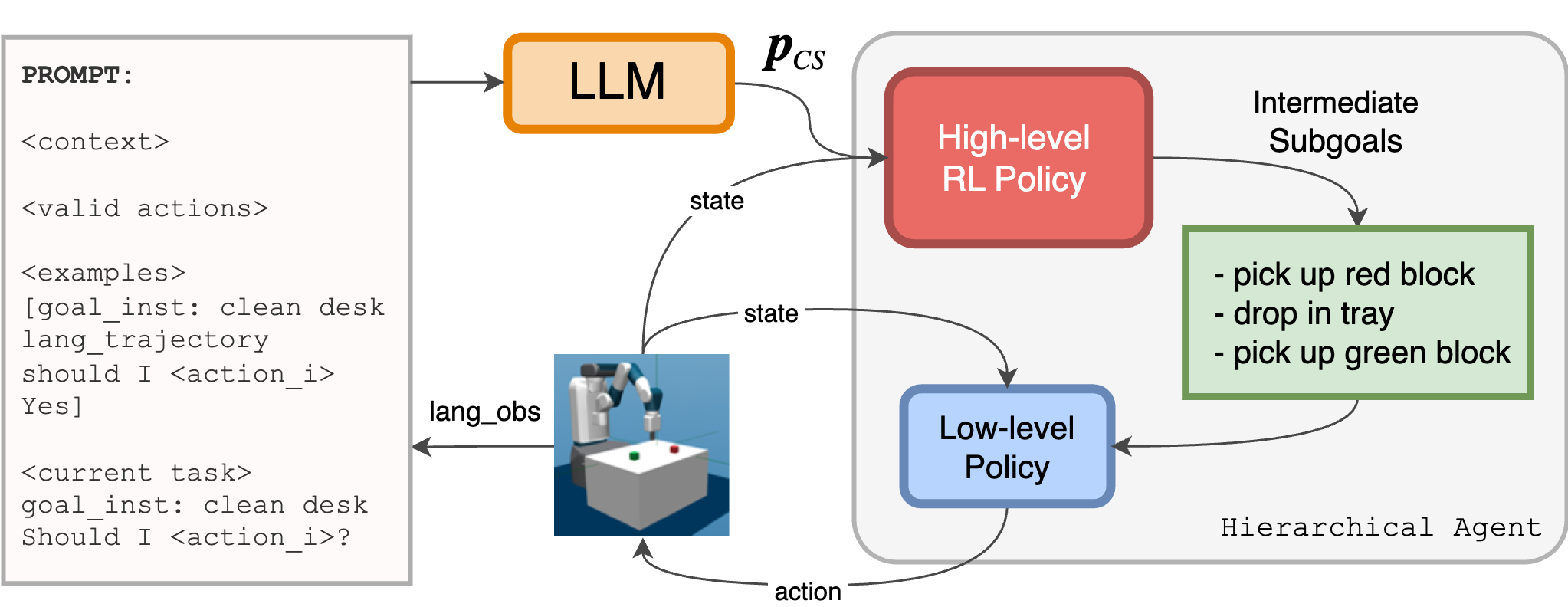}
  \caption{The LLM to guides the high-level policy and accelerates learning. It is prompted with the context, some examples, and the current task and observation. The LLM's output biases high-level action selection}
  \label{fig:arch}
  \vspace*{-0.2cm}
\end{figure}

One of the reasons that humans are so good at dealing with unfamiliar situations is that we almost never solve problems from scratch. Presented with a new task and a library of skills, we are able to choose a subset of skills that seem most relevant and explore from there. We might perform some trial and error exploration (as in RL), but we quickly learn the right subset of skills as well as the correct sequence in which they need to be executed. 
For example, the door handles on newer cars lie flat against the door, unlike most other car door handles in existence. That presents a problem the first time you try to open one. Humans immediately narrow down to a few exploratory actions, like trying to get a finger under the handle or pushing on it in different places.  We don't, unlike most RL algorithms might, tap the window or pull the side mirrors, as we believe that such options are causally irrelevant based on deep world knowledge.

Large language models (LLMs) have been shown to encode a tremendous amount of knowledge about the world by virtue of being trained on massive amounts of text.  We hypothesize that this knowledge can be leveraged to focus the training of hierarchical policies, making them significantly more sample efficient.
In particular, we explore how large pretrained language models can be used to inject common sense priors into hierarchical agents. 

In this approach we assume access to several low level skills. These can be, for example, engineered planners or policies learned using RL and sub-task rewards. Based on a high-level task description and current state, the LLMs guide the agent by suggesting the most likely courses of action. Instead of random exploration, we use these suggestions to intelligently explore the various options. Because LLMs are not grounded in the domain, they are only used to bias action selection and their influence is reduced as training progresses. This results in a policy that can be deployed without depending on the LLM at run time. We evaluate this approach on several simulated environments (MiniGrid \cite{gym_minigrid}, SkillHack \cite{matthews2022hierarchical}, and Crafter \cite{hafner2021crafter}),  showing that it can learn to solve complex, long-horizon tasks much faster than baseline methods. Experiments with a real robot arm in block manipulation tasks using a tabular Q-learning version of the same algorithm show that it can learn policies much faster with less experience in that domain. Our contributions are summarized as follows:
\begin{itemize}
    \item an approach for using LLMs to guide exploration by extracting common sense priors;
    \item a hierarchical agent that uses these priors to solve long-horizon tasks;
    \item an evaluation of the framework in simulation as well as a simple real-world environment, showing that it performs significantly better than baselines;
    \item a discussion of (1) the advantages of our method compared prior work and (2) potential future work.
\end{itemize}

\section{Related Work}

\textbf{Langauge and HRL} There is significant prior work on hierarchical RL where the standard MDP is converted into a semi-Markov decision process (SMDP). The most common approach is to incorporate temporally extended actions, also known as options or skills \cite{bacon2017option}. Typically, a low-level policy achieves sub-tasks by executing primitive actions and a high-level policy plans over temporally extended options or skills. Natural language is a popular way to specify sub-tasks and achieve generalization due to its inherent compositionality and hierarchical structure \cite{jiang2019language, prakash2021interactive, waytowich2019b, hu2019hierarchical}. Most of these methods specify or generate a high-level plan in natural language, which is then executed sequentially by a separate low-level policy. These approaches face challenges when operating in high-dimensional observation spaces. They also rely on manual data collection to train the high-level policies and are therefore difficult to generalize to new tasks \cite{andreas2017modular, mirchandani2021ella, goyal2019using}.

\textbf{RL and Foundation Models} Recently, large language models such as GPT-3 have been used to build agents capable of acting in the real world based on language instructions \cite{brown2020language, bommasani2021opportunities}. The in-context learning and intelligent prompting strategies supported by these models have been used to design language-guided hierarchical agents. \cite{huang2022language} use LLMs as zero-shot planners to enable embodied agents to act in real world scenarios. Similarly, \cite{ahn2022can} use LLMs along with affordance functions to generate feasible plans that guide a robot to achieve goals specified in natural language instructions.
Our work is closely related to \cite{du2023guiding},  where they improve exploration by using LLMs to provide intermediate rewards and encourage the agent to seek novel states.


\section{Methods}

\subsection{Problem Statement}
We consider a system that receives instructions in the form of natural language describing a task, similar to \cite{ahn2022can}. The instructions can be long, may contain warnings and constraints, and may not include all of the necessary individual steps. We also assume that the agent has access to a finite set of skills or sub-policies that can be executed in sequence to solve long-horizon tasks. These skills can be hand-coded, or trained using reinforcement learning or imitation learning with manual reward design.
They must be accompanied by a simple description in natural language, such as "pick up red block" or "open blue door". They must also be able to detect sub-task completion to switch control back to the high-level policy. Given a finite set of options or skills, our objective is to obtain a high-level decision policy that selects among them.

\subsection{Using LLMs to Guide High-level Policies}

This section introduces our method for using LLMs to improve exploration in the high-level policy of an HRL system. The semantic knowledge and planning capabilities of LLMs  improve high-level action selection given a task description and current state in the form of language.
The core idea is to use LLMs to obtain a value that approximates the probability that a given skill or sub-task is relevant to achieve the larger goal. As mentioned earlier, each skill is accompanied by a language description $l_{skill}$ and the current trajectory is translated into language, $l_{traj}$. There is also  a high level instruction, $l_{goal\_inst}$, describing the larger goal along with optional constraints. 

The LLM is used to evaluate the function $f_{LLM}(l_{skill_i}, l_{goal\_inst}, l_{traj})$ for each skill at every high-level decision step. Essentially, the LLM  answers the following question: given the task, $l_{goal\_inst}$, and trajectory so far, $l_{traj}$, should we choose skill $l_{skill_i}$? The output of the LLM, `yes' or `no', can easily be converted to an int (``0'' or ``1''). This kind of closed form question-answering prompt has been shown to work better than open ended prompts \cite{du2023guiding}. After evaluating this for each of the $k$ skills, we get $ F_{LLM} = [f_{LLM_1}, f_{LLM_2}, f_{LLM_3}, ... ,f_{LLM_k}]$. For example, $F_{LLM} = [0, 1, 0,..., 0, 1, 0, 0] $.
A {\tt LOG SOFTMAX } function is applied to these logits to get the common sense priors from the LLM denoted by $p_{CS} = log\_softmax(F_{LLM})$. 
Relying entirely on $p_{CS}$ is not enough to solve complex tasks. At the same time, using RL and exploring without any common sense intuition is inefficient. Therefore we still use RL and sparse rewards to obtain high-level policies but also use the common sense priors, $p_{CS}$, from the LLM to guide  exploration. More details about the RL algorithms used are in the Experiments section. The action selection in the exploration policy samples actions from a categorical distribution where the logits are obtained by the policy head processing the state. These logits are biased with the common sense priors $p_{CS}$ and a weight factor $\lambda$. So the action selection looks like this: $ a = {\tt Categorical}[\pi(s_t) + \lambda.p_{CS}(s_t)] $. Here, the action $a$ is the temporally extended macro action or the skill. The weight factor starts from $\lambda = 1$ and is annealed gradually until it reaches zero by the end of training. This means that the trained agent does not continue to reply on the LLM during deployment.
This process is summarized in Algorithm \ref{alg:algo1} and Figure \ref{fig:arch}.

\begin{algorithm}[t]
  \caption{}
  \label{alg:algo1}
  \begin{algorithmic}
    \State ${\tt high\_inst } \gets$ high level goal in language
    \State ${\tt \pi_\theta } \gets$ high level policy
    \State $f_{LLM} \gets$ common sense priors from LLM 
    \Procedure{ {\tt LLMxHRL(high\_inst)}}{}
    \State init  ${\tt \pi_\theta }$
    \While{${\tt \pi_\theta }$ not converged}
    \State init $\tau \gets \{\} $ 
    \For{$t \gets 0$ \textbf{to} $T$}
    \State $p_{CS} \gets  f_{LLM}(high\_inst, \tau)$ 
    \State $a_t \gets cat\_dist[\pi_\theta(\tau)+\lambda.p_{CS}(\tau)].sample()$
    \State $s_t, r_t \gets act(a_t)$
    \State $ \tau \gets append(s_t, r_t, a_t)$
    \EndFor
    \State update $\pi_\theta$ 
    \EndWhile
    \State return $\pi_\theta$ 
    \EndProcedure
  \end{algorithmic}
\end{algorithm}
\vspace*{0.2cm}


\textbf{LLM Queries and Prompt Design.} We use the \emph{gpt-3.5-turbo} GPT provided by OpenAI APIs.  To reduce the number of API calls, the LLM responses for all possible combinations of $l_{goal\_inst}$ and $l_{traj}$ are cached. A simplified version of $l_{traj}$ is used to denote the current trajectory history using the past two actions. The main prompt used in our experiments has the following structure
{\tt Goal:$l_{goal\_inst}$, So far I have: $l_{traj}$, Should I $l_{skill_i}$? }. The LLM is shown a few examples of responses to such queries and the prompt specifies that a one word Yes/No answer is required. Example prompts are in the Appendix.


\section{Experiments}

This section describes the experimental setup and results of testing the framework in three simulation environments and one real world robotic arm block manipulation task. The framework relies on communicating with the LLM using text. As mentioned earlier, each skill corresponds to a text description $l_{skill_i}$ and the larger goal is described using $l_{goal\_inst}$. We assume access to a captioner that maps the current observation history to $l_{traj}$. This could be automated by using modern vision to language models such as \cite{radford2021learning}, but that is left for future work. Instead we use a CLIP-based model along with an LLM in the experiments with a real robot to convert visual input to a discrete low dimensional state. More details about obtaining $l_{traj}$ are in the Appendix. In each environment,  our method is compared with baseline hierarchical agents without any guidance from LLMs, and an oracle and a SayCan-like agent without affordances.

\begin{table}[!b]
\small
\centering
\begin{tabular}{p{2.5cm}|p{9.5cm}}
\toprule
\textbf{Method} & \textbf{Description} \\ \midrule
LLM x HRL (ours)    &  Use an LLM to bias high-level action selection as explained in Section 3. Only receives reward on task completion. \\  \midrule
Vanilla HRL   &  A baseline hierarchical agent that has no guidance from the LLM.  \\  \midrule
Shaped HRL   &  Same as the Vanilla HRL with no LLM guidance. But here agents receive shaped rewards for successful sub-task completion. Requires hand-engineered reward functions.   \\  \midrule
Oracle  &  This is the upper bound. The high-level policy is an oracle state-machine that provides the right sub-tasks in the correct sequence. \cite{goecks2021combining} \\  \midrule
SayCan w/o Aff  &  A SayCan \cite{ahn2022can} like architecure but without an affordance function that blindly trusts the LLM. This method requires LLM access during deployment\\  \midrule
\end{tabular}
\vspace*{0.2cm}
\caption{A description of the methods used in the experiments} 
  \label{comp_table}
\end{table}




\subsection{MiniGrid Experiments}

\textbf{Setup} The experiments described in this section were performed on the MiniGrid environment by \cite{gym_minigrid}, which is a simple grid world. The environment can be designed with multiple rooms with doors, walls, and goal objects. These objects can have different colors and the agent and goal objects are spawned at random locations.
The action space is discrete which allows movement in the
4 compass directions, opening and closing doors, and picking up and
dropping objects. We designed multiple tasks in this setup which can be broken down into smaller sub-tasks. 
\begin{itemize}
    \item The \emph{UnlockReach} task consists of a random object in a room which are behind a locked door. The agent has to first find the right key based on the door color, unlock the door, and then navigate to the goal object. 
    \item The \emph{KeyCorridor v0} task consists of a corridor with multiple rooms on either side. A goal object is inside a locked room whose key is in another room. The agent has to first find the key and then unlock the door to ultimately reach the goal
    \item \emph{KeyCorridor v1} is similar to v0, but some of the rooms have defective keys. The goal instruction comes with the rooms to avoid. This task is much more difficult for standard HRL methods.
\end{itemize}

\begin{figure}[t]
  \centering
  \begin{subfigure}{0.32\textwidth}
    \includegraphics[width=\linewidth]{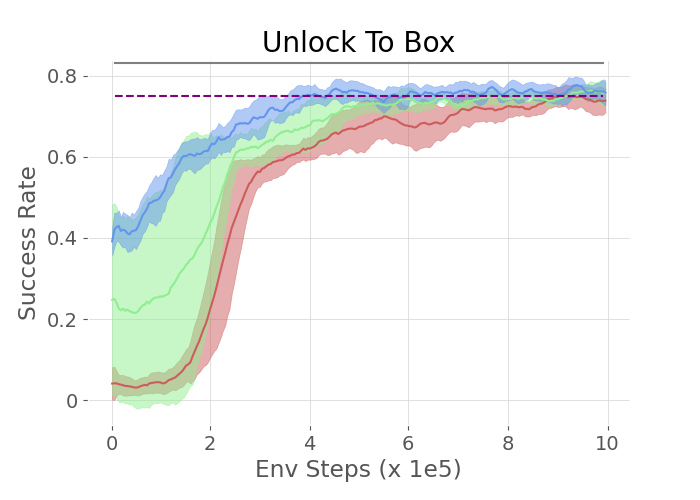}
  \end{subfigure}
  \hfill
  \begin{subfigure}{0.32\textwidth}
    \includegraphics[width=\linewidth]{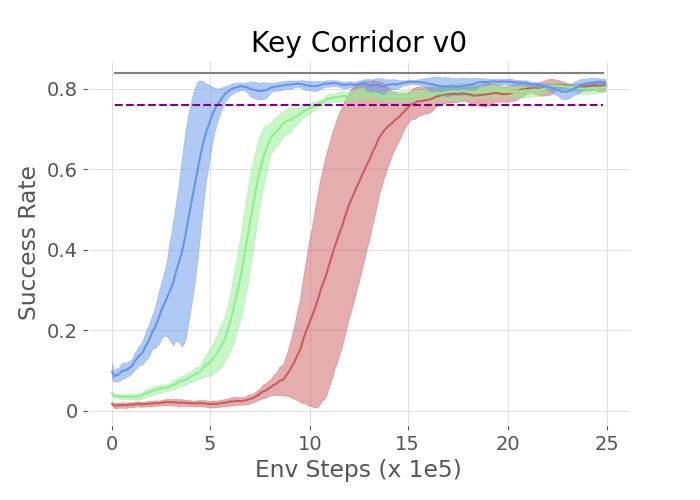}
  \end{subfigure}
  \hfill
  \begin{subfigure}{0.32\textwidth}
    \includegraphics[width=\linewidth]{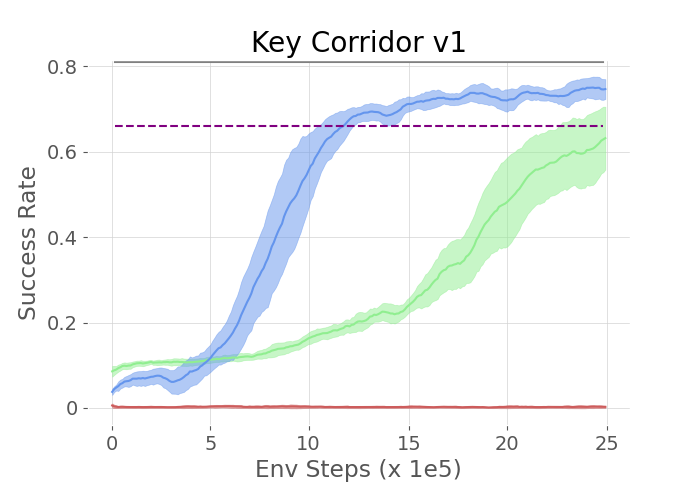}
  \end{subfigure}

 \begin{subfigure}{0.9\textwidth}
    \includegraphics[width=\linewidth]{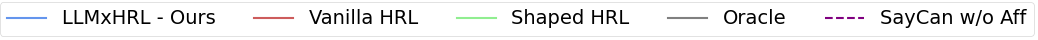}
  \end{subfigure}
  
  \caption{The plots show the success rate of different methods on the three tasks in the MiniGrid Environment.}
  \label{fig:MiniPlots}
\end{figure}

Each task has a single reward that is only provided on successful task completion. The agents have access to several temporally extended skills: \emph{GoToObject, PickupObject, UnlockDoor}, and \emph{OpenBox}. These are conditioned on the type and color of the objects. For example the \emph{Object} may refer to a \emph{key, ball}, or \emph{box}, and the color can be \emph{red, green, blue, yellow}, etc. These low-level sub-tasks were pretrained and frozen using PPO \cite{schulman2017proximal} and manual reward specification. The high-level policies are also trained using PPO where the sub-tasks are treated as actions. We compared against Vanilla HRL, Shaped HRL, an Oracle, and a SayCan-like method as described in Table \ref{comp_table}. The results are summarized in Figure \ref{fig:MiniPlots}. It's clear that our method outperforms both the baseline HRL methods with and without shaped rewards. It is also able to converge to the optimal policy much sooner than the other methods. The Oracle and SayCan are not trained using RL and so we show their performance using horizontal lines. Although they are comparable to our method, one benefit of our method is that it does not rely on the LLM during deployment.

\subsection{SkillHack}

The NetHack Learning Environment \cite{kuttler2020nethack} is an RL environment based on the classic game of NetHack. It is notoriously difficult because of the large number on entities and actions, procedural generation, and the stochastic nature of the game. MiniHack \cite{samvelyan2021minihack} and SkillHack \cite{matthews2022hierarchical} are extensions of NetHack that enable creation of custom levels and tasks.  They are simpler than the full game while retaining most of the interesting complexities. The SkillHack suite contains 16 skills such as \emph{PickUp, Navigate, Fight, Wear, Weild, Zap, Apply}, etc. More details are in the Appendix. These skills can be executed sequentially to achieve larger tasks. We consider two such tasks - \emph{Battle, FrozenLavaCross}. 

\begin{itemize}
    \item In the \emph{Battle} task, the agent needs to \emph{PickUp} a randomly placed Sword, \emph{Wield} the Sword and finally \emph{Fight} and kill a Monster. 
    \item In the \emph{FrozenLavaCross} task, the agent needs to \emph{PickUp} either a WandOfCold or a FrostHorn based on what is available, then create a bridge across the lava with either \emph{ZapWandOfCold} or \emph{ApplyFrostHorn}. Finally, the agent must \emph{NavigateLava} across the newly made bridge to reach the staircase on the other side.
\end{itemize}

\begin{figure}[t]
  \centering
  \begin{subfigure}{0.24\textwidth}
    \includegraphics[width=\linewidth]{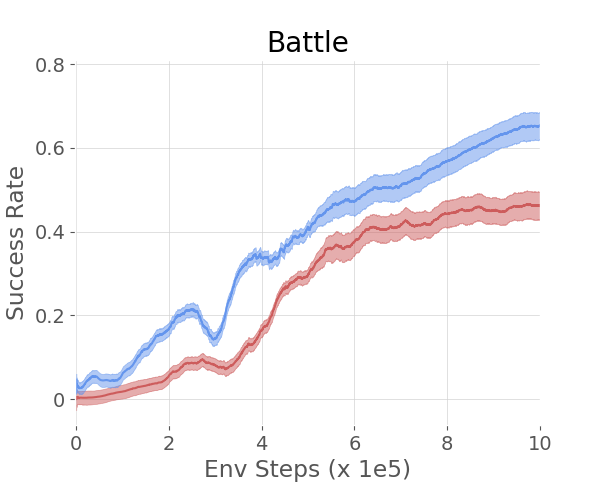}
  \end{subfigure}
  \hfill
  \begin{subfigure}{0.24\textwidth}
    \includegraphics[width=\linewidth]{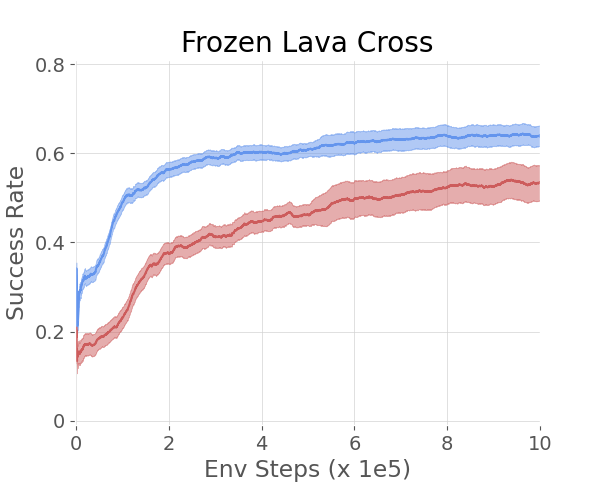}
  \end{subfigure}
  \hfill
  \begin{subfigure}{0.24\textwidth}
    \includegraphics[width=\linewidth]{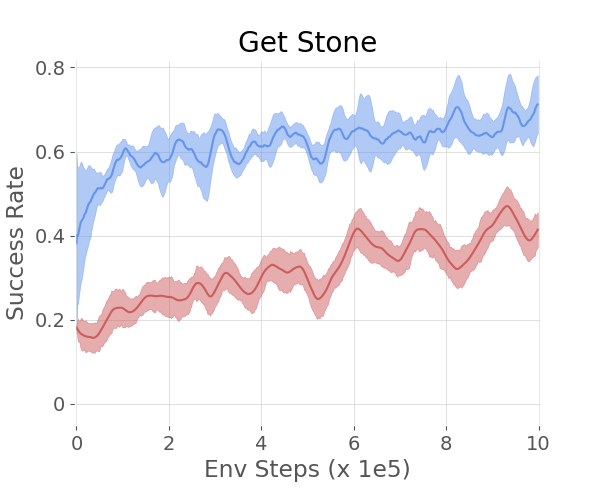}
  \end{subfigure}
  \begin{subfigure}{0.24\textwidth}
    \includegraphics[width=\linewidth]{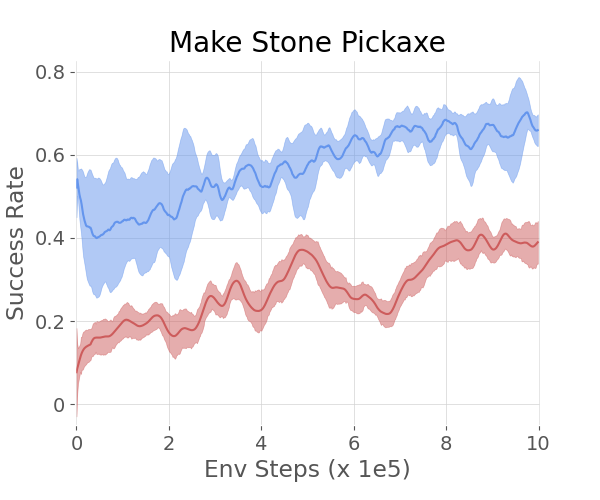}
  \end{subfigure}
  
  \begin{subfigure}{0.4\textwidth}
    \includegraphics[width=\linewidth]{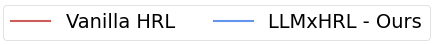}
  \end{subfigure}

  \caption{The 2 plots on the left show the success rate of different methods on the SkillHack - Battle and Frozen Lava Cross. The 2 plots on the right show the success rate of different methods on the Crafter - Get Stone and Make Stone Pickaxe}
  \label{fig:skill_craft_plots}
  \vspace*{-0.4cm}
\end{figure}

In this environment we compare against Vanilla HRL and an Oracle high-level policy. The low level skills are trained using IMPALA \cite{espeholt2018impala} with the code provided by \cite{matthews2022hierarchical}. The high-level policy is also trained using IMPALA where the policy skills are macro actions.
As seen in the first two plots in Figure \ref{fig:skill_craft_plots}, in both the tasks, \emph{Battle} and \emph{FrozenLavaCross}, our method clearly outperforms the HRL agent without LLM guidance.

\subsection{Crafter}

Crafter \cite{hafner2021crafter} is a 2D version of Minecraft which has the same complex dynamics but with a simpler observation space and faster simulation speeds. Similar to Minecraft, it involves collecting and building artifacts along an achievement tree. We modified the game slightly to make it easier by slowing down health degradation and having fewer dangers to fight. We evaluated on two tasks that have a natural hierarchical structure - \emph{MakeWoodPickaxe} and \emph{MakeStonePickaxe}. More details are in the Appendix. 
Similar to our other experiments we pretrain policies for multiple skills using PPO. The high level policy is then trained to select among these skills. The last two plots in Figure \ref{fig:skill_craft_plots} show how our method performs better than the baseline HRL method.

\subsection{uArm Real Robot Experiments}

We also tested on a real robot arm on a simpler tabular Q-learning version of our method. uArm Swift Pro \cite{uarm-sdk} is an open-source desktop robotic arm. 
\begin{wrapfigure}{l}[0pt]{0.5\textwidth}
\vspace*{-0.5cm}
  \begin{center}
    \includegraphics[width=0.4\textwidth]{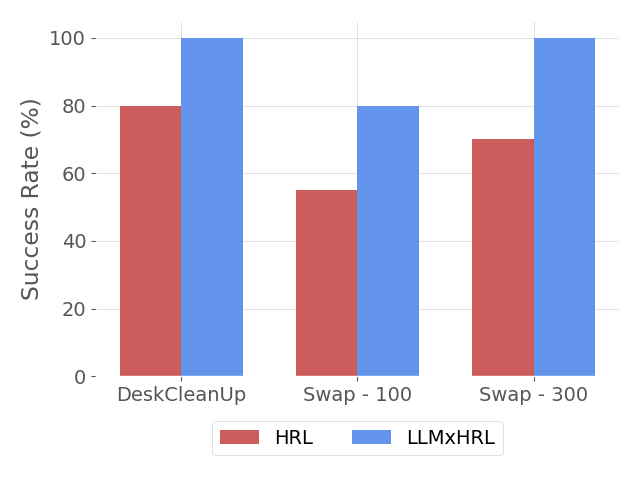}
  \end{center}
  \vspace*{-0.3cm}
  \caption{Robot Arm Results}
  \vspace*{-0.4cm}
\label{fig:uarmPlots}
\end{wrapfigure}
We designed two block manipulation tasks - \emph{DeskCleanUp} and \emph{SwapBlocks}. Similar to the previous simulation experiments, we assume access to various skills that can be used to solve larger, more complex tasks. In our setup, video from a camera is used to convert the robot arm and block positions into an array of discrete values representing the state. From this simplified state, we are able to learn a high-level policy with tabular Q-learning. Like before, we calculate $p_{CS} = log\_softmax(F_{LLM})$ using an LLM and access to $f_{LLM}(l_{skill_i}, l_{goal\_inst}, l_{traj})$. Refer to the Appendix for more details.

Figure \ref{fig:uarmPlots} show the results of our experiments. In the \emph{DeskCleanUp} task, there are three locations with a tray and two blocks (red and green). The episode is initialized with blocks in  random locations. The goal is to pick up the blocks and place them in the tray, essentially cleaning the desk. This task was trained for 100 episodes. In the \emph{SwapBlocks} task there are also three locations (or zones) with two blocks in two random locations. The goal is to swap the positions of the blocks. In  Figure \ref{fig:uarmPlots},  \emph{Swap - 100} denotes performance after 100 episodes and \emph{Swap - 300} is after 300 episodes. We can see that using LLMs to guide agent exploration give us better performance in fewer trials.

\section{Discussion}
In this work we present a framework for using LLMs to guide exploration in hierarchical agents. Instead of learning from random exploration without any prior knowledge, we use the LLMs to suggest high-level actions based on the task and current state. We evaluate our method on long horizon tasks in simulation environments as well as with a real robot. We show that out method performs better than baselines and does not require manual reward shaping. Moreover, once the agent is trained, it no longer depends on the LLM during deployment unlike some prior methods. 

This work can be extended in several ways to make it more end-to-end. We Currently assume access to a function that provides us  language descriptions of the current trajectory and state. This can be automated using recent advancements in vision language models (VLM). It will also be interesting to extend this framework for more than one level or hierarchy to tackle longer tasks.

\section{Acknowledgments}

This project was sponsored by the U.S. Army Research Laboratory under Cooperative Agreement Number W911NF2120076.

\bibliographystyle{abbrvnat}
\bibliography{rl.bib}

\newpage

\appendix

\section{Appendix}
\subsection{Environment Details}
\subsubsection{MiniGrid}
\begin{figure}[H]
  \centering
  \begin{subfigure}{0.27\textwidth}
    \includegraphics[width=\linewidth]{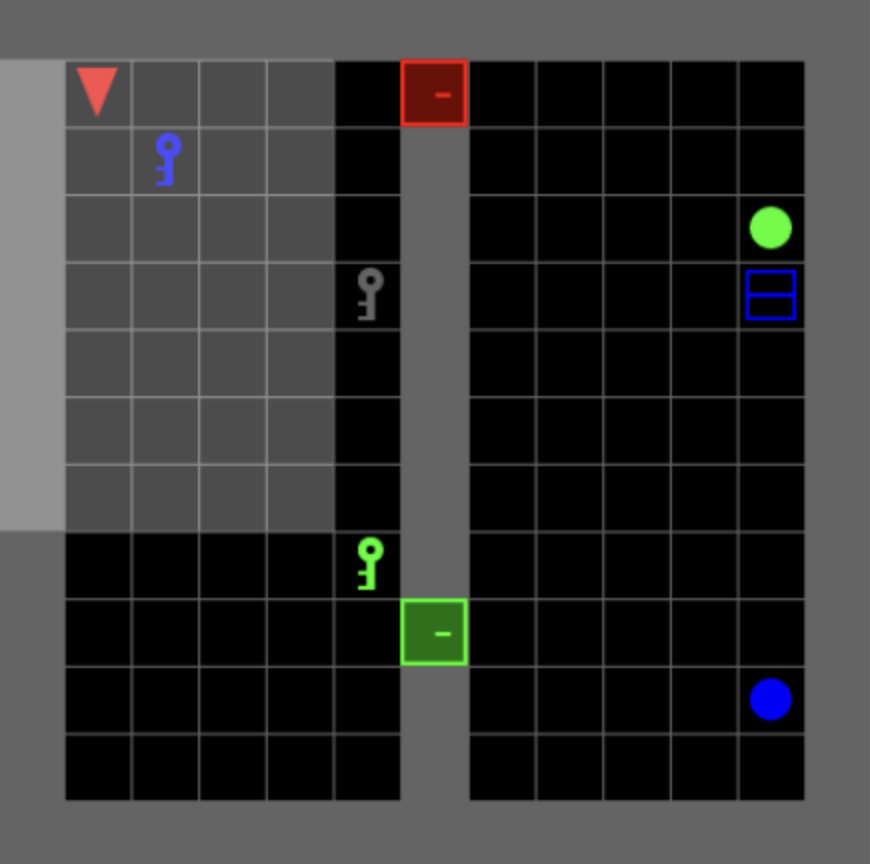}
    \caption{Goal: open the locked green door and go to the blue box \newline}
  \end{subfigure}
  \hspace*{0.5cm}
  \begin{subfigure}{0.27\textwidth}
    \includegraphics[width=\linewidth]{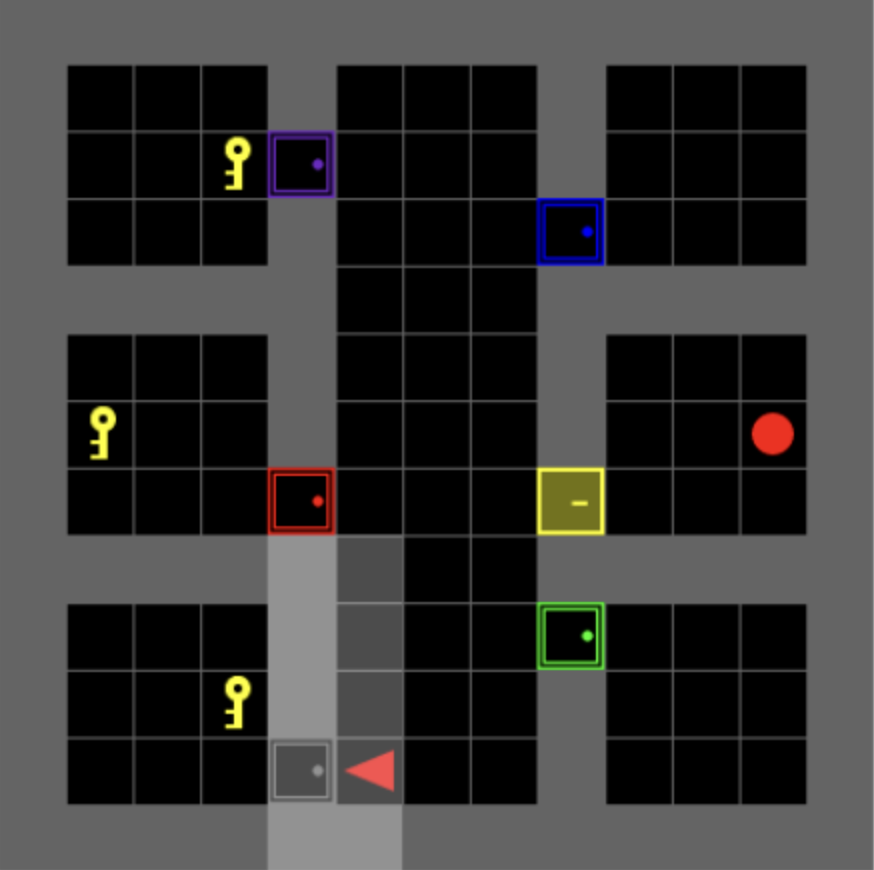}
    \caption{Goal: pick up a purple key, then open the purple door and go to the red ball \newline}
  \end{subfigure}
  \hspace*{0.5cm}
  \begin{subfigure}{0.27\textwidth}
    \includegraphics[width=\linewidth]{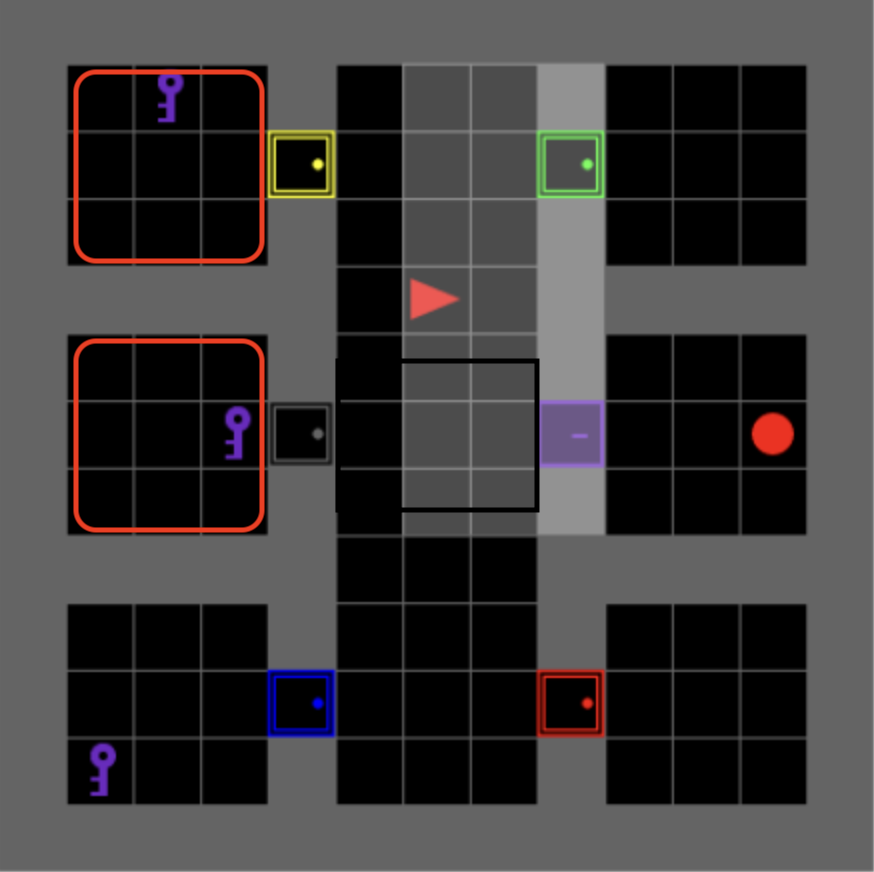}
    \caption{Goal: pick up a purple key, then open the purple door and go to the red ball, avoid the yellow and grey doors}
  \end{subfigure}

  \caption{The agent is represented using the red triangle. Left: The UnlockReach task where the agent needs to get the right key and open a door and then go to the object in the right room. Middle: The KeyCorridor-v0 task where the agent needs to reach the red ball in one of the locked rooms on the right. It first needs to get a the key to from one of the rooms on the left. Right: Similar to v0 but the some of the rooms have defective keys shown in red. The agent does not see this, it only recieves this information in the text goal}
  \label{fig:mg_envs}

\end{figure}

Minigrid is an open source gridworld environment \cite{gym_minigrid}. We use three tasks  \emph{UnlockReach, KeyCorridor v0 and KeyCorridor v1}.
Figure \ref{fig:mg_envs} shows the grid layouts for the three tasks. The observation is an encoded version of the grid which capture the each cell type, color and an optional door/box state. We consider the fully observable version of these tasks, which means the observations consists of the full grid - 13x13 in our case.

\subsubsection{SkillHack}
SkillHack \cite{matthews2022hierarchical} is an extensions on top of \cite{kuttler2020nethack} which where you can design custom levels and get visual/spacial states along with text descriptions. Figure \ref{fig:skill_craft_envs} shows the tasks we test where each of them require solving multiple sub-tasks. The state consists of a 2D map along with inventory information and text describing the effect of each action. This is very convenient for out method as we need to interact with the LLM using language.

\subsubsection{Crafter}
\begin{figure}[H]
  \centering
  \begin{subfigure}{0.32\textwidth}
    \includegraphics[width=\linewidth]{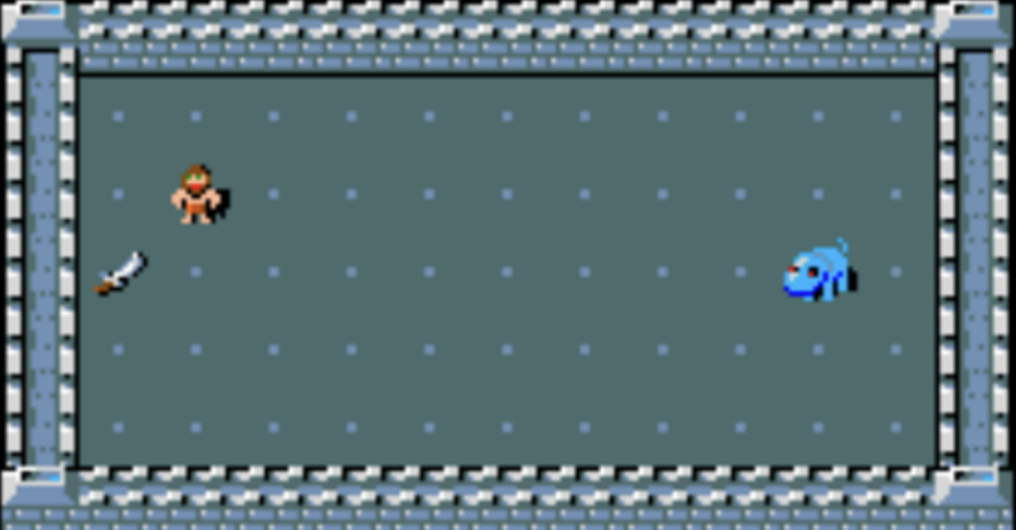}
    \caption{Battle: The goal is to pick uo the sword, navigate to the monster and fight \newline}
  \end{subfigure}
  \hspace*{0.2cm}
  \begin{subfigure}{0.33\textwidth}
    \includegraphics[width=\linewidth]{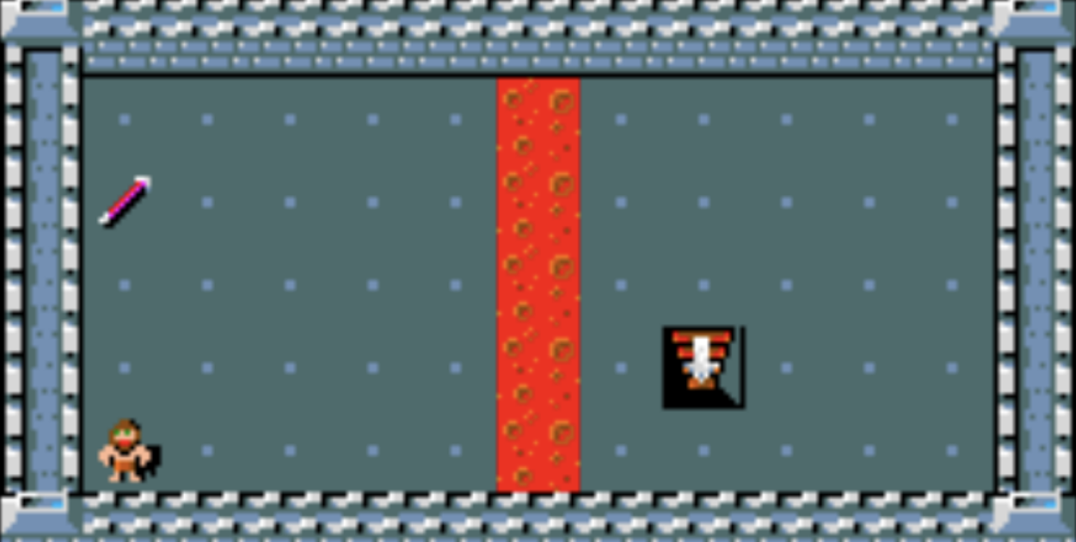}
    \caption{FrozenLavaCross: Here the goal is to pick up the wand, navigate to the lava to zap, creating a bridge and then navigate to the exit}
  \end{subfigure}
  \hspace*{0.5cm}
  \begin{subfigure}{0.2\textwidth}
    \includegraphics[width=\linewidth]{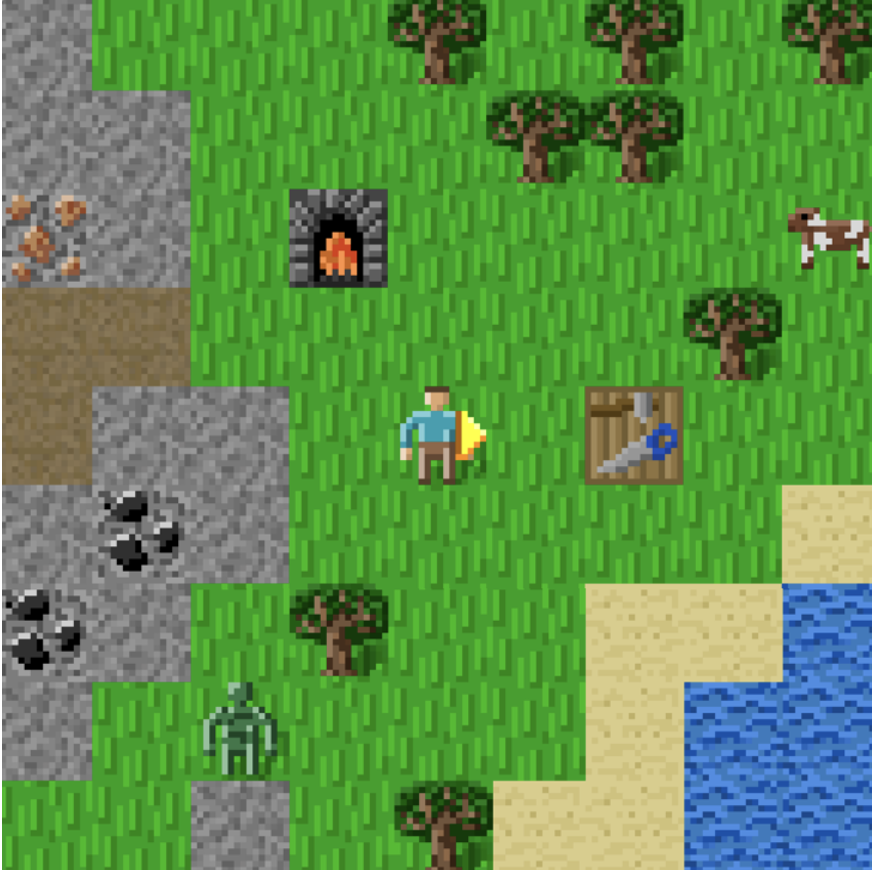}
    \caption{Crafter Env: 2D Minecraft like environment}
  \end{subfigure}
  
  \caption{Left and Middle: The SkillHack environment suite based on the text based NetHack game. Right: Crafter: Here, the agent explores the open world environment to collect resources like wood, stone and craete new objects and tools}
  \label{fig:skill_craft_envs}
\end{figure}

Crafter \cite{hafner2021crafter} is a 2D version Minecraft, Figure \ref{fig:skill_craft_envs}. it has a very simple state representation encoding the items on the map, an inventory and the health of the agent. It is fairly easy to translate this into text using a hand-coded function. The high-level skills $l_{skill_i}$ we consider, such as \emph{chop tree, create table, make wood pickaxe} etc can also be naturally described using langauge phrases.

\begin{figure}[H]
  \centering
  \begin{subfigure}{0.27\textwidth}
    \includegraphics[width=\linewidth]{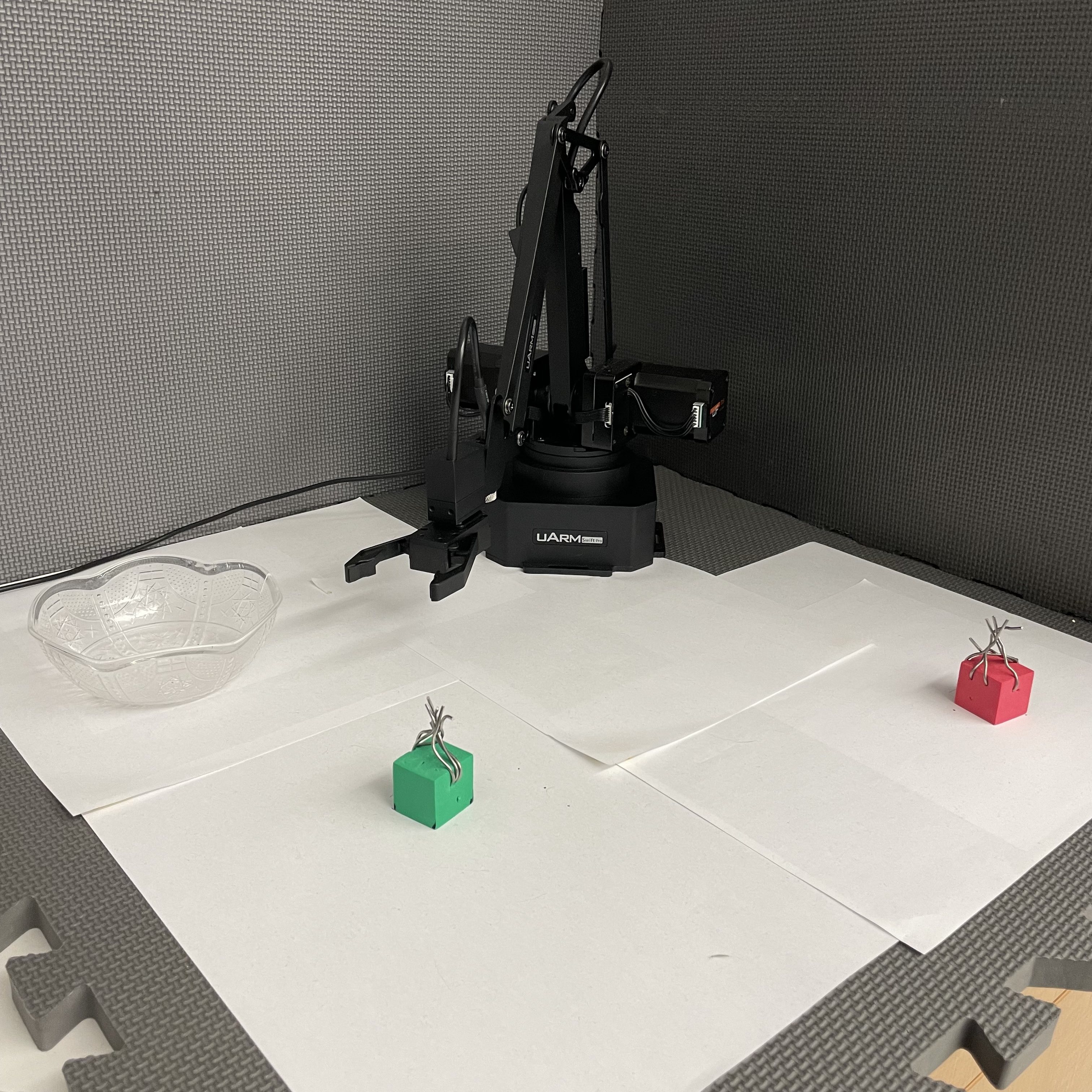}
    \caption{Desk Cleanup Task}
  \end{subfigure}
  \hspace*{0.5cm}
  \begin{subfigure}{0.27\textwidth}
    \includegraphics[width=\linewidth]{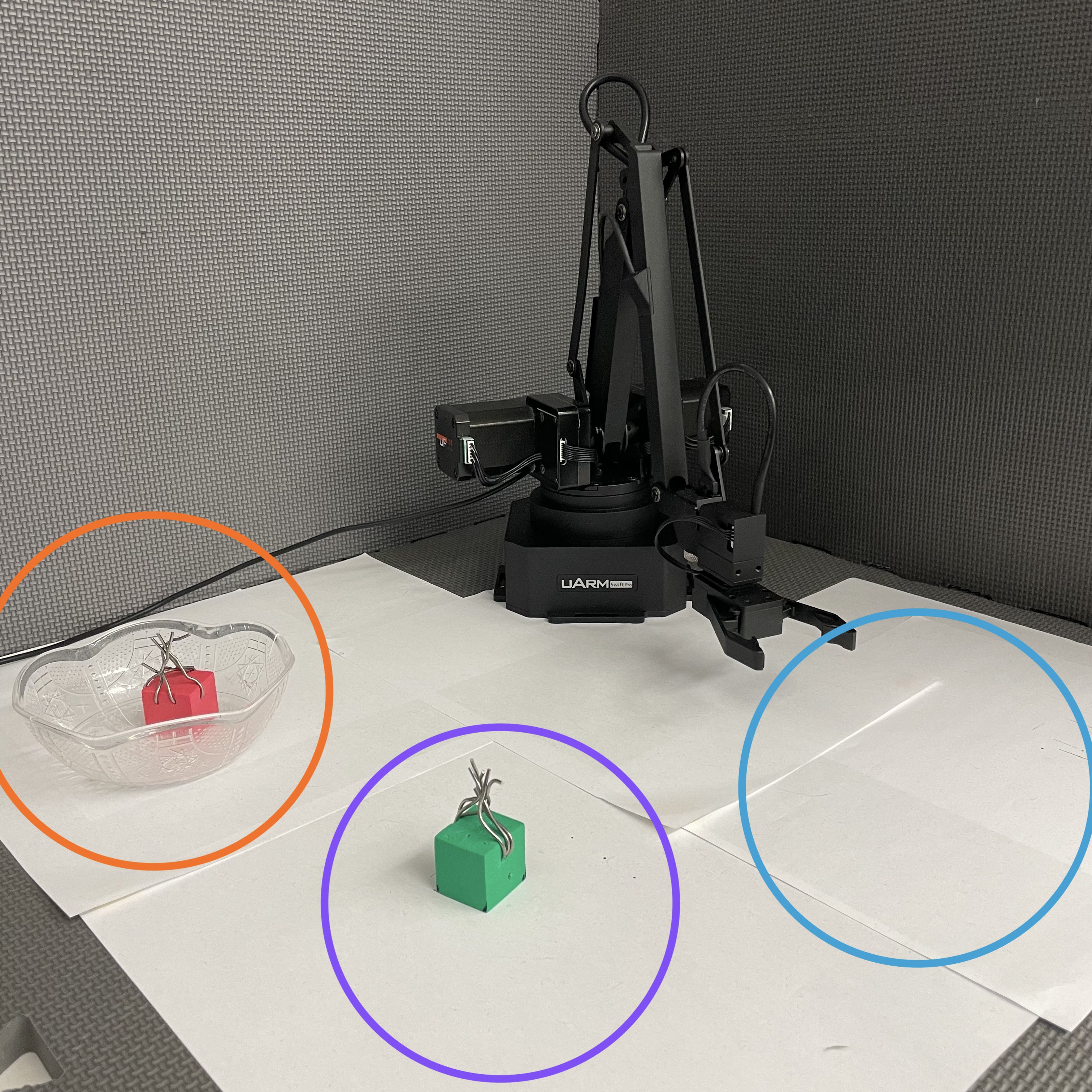}
       \caption{Desk Cleanup Task}
  \end{subfigure}
  \hspace*{0.5cm}
  \begin{subfigure}{0.27\textwidth}
    \includegraphics[width=\linewidth]{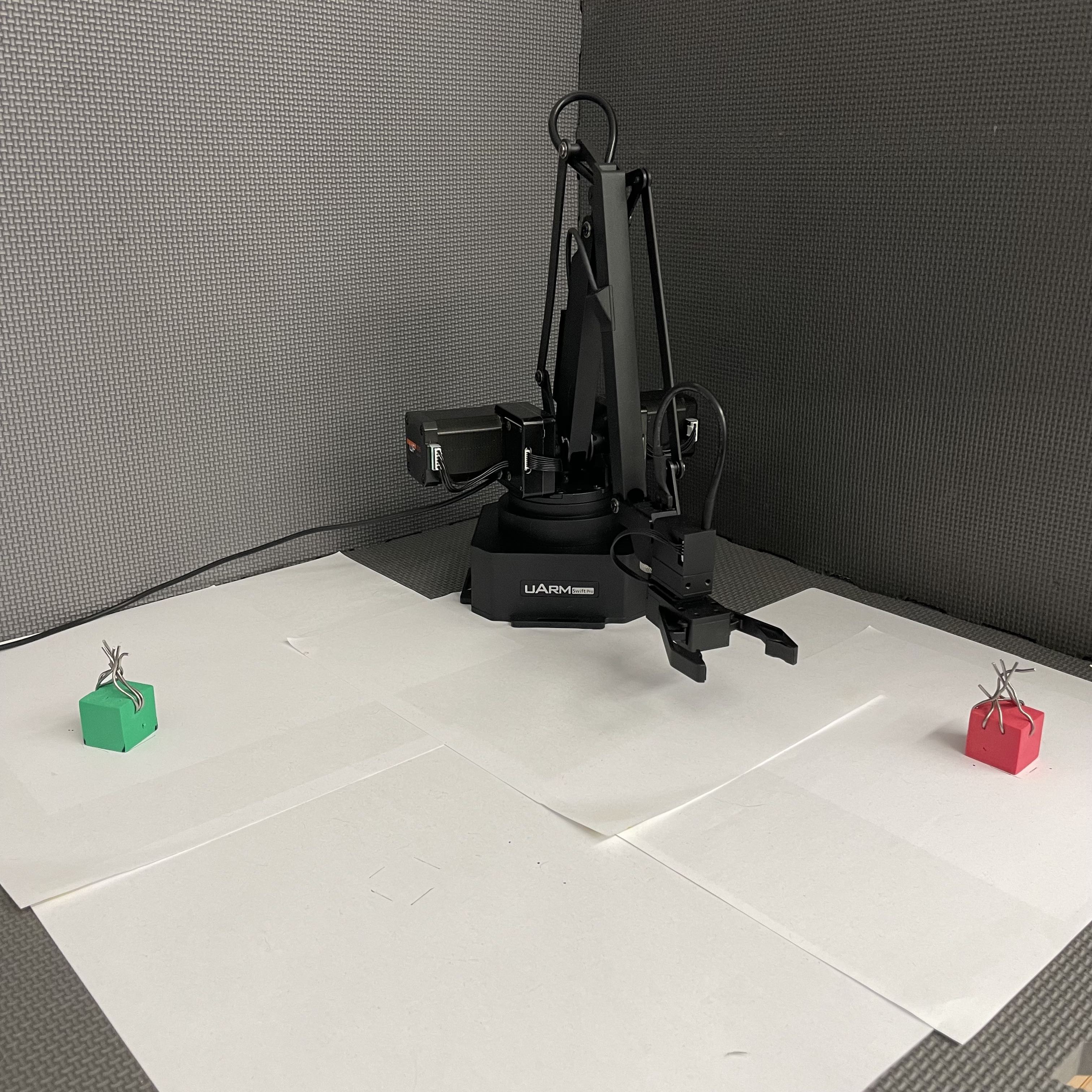}
    \caption{Block Swap Task}
  \end{subfigure}

  \caption{uArm: Robot arm environment}
  \label{fig:uarm_env}

\end{figure}

\subsubsection{uArm Robot Arm}

uArm Swift Pro is an open source research robotic arm. We use this to test the tabular Q-leraning version of our method using two tasks \cite{uarm-sdk}. The environment is setup where the table has 3 zones as shown in Figure \ref{fig:uarm_env} (b). The robot is able to pick and place objects from any of these zones. They are considered the low level skills and are hard-coded. However with more resources, these can be trained using RL making them more robust. 
In the \emph{DeskCleanup} task, there is a tray in one of the zones as seen in Figure \ref{fig:uarm_env} (a) and (b). The goal is to pick up the blocks and place them in the tray.
In the \emph{BlockSwap} task the two blocks are placed in random zones Figure \ref{fig:uarm_env} (c). The goal is to swap the positions. This can only be done by utilizing the empty zone. We use an Intel Realsense camera to convert the visual information to a simplified discrete state for the tabular Q-learning. The state consists of 4 discrete values, $[arm\_location, holding, red\_location, green\_location]$. We get the $arm\_location$ form the robot apis, which is one of the 3 zones. The $red\_location$ and $green\_location$ denote the location of the red and green blocks respectively (one of 3 zones). $holding$ denotes the block the arm is currently holding if any. We extract $holding, red\_location$ and 
$green\_location$ from the camera image. Instead of training or finetuning a separate  object detection model we use CLIP and text phrases describing the objects on image patches. We then get the co-ordinates and map them to the right values in the state space.

\subsection{Prompt Design}
We use the following structure for all out prompts. 
\newline \newline 
{\tt Goal:$l_{goal\_inst}$ \newline 
So far: $l_{traj}$ \newline  Should I $l_{skill_i}$? \newline  }
\newline 
$l_{traj}$ represents the trajectory history which captures what the agent has done so far. We find that the 2 actions are sufficient to capture this as that the tasks we test on not extremely complicated. As mentioned in previous sections, $l_{goal\_inst}$ describes the goal along with details, warnings and constraints. Each skill is associated with a language description $l_{skill_i}$.

\subsubsection{MiniGrid Prompts}

\begin{lstlisting}[gobble=0, language=plantuml, breakautoindent=false, frame=trBL]
You are a 2D maze-solving agent with access to a variety of low-level skills such as "pick:red:ball", "pick:red:key", "pick:green:ball", "pick:green:key", "pick:blue:ball"...

Goal: open the locked green door and go to the blue box
So far: 
Question: Should I pick:red:ball?
Answer: No

Goal: open the locked green door and go to the blue box
So far: 
Question: Should I  goto:blue:box?
Answer: No

Goal: open the locked green door and go to the blue box
So far:  pick:green:key
Question: Should I unlock:green:door?
Answer: Yes

Goal: open the locked green door and go to the blue box
So far:  pick:green:key, unlock:green:door
Question: Should I  goto:red:box?
Answer: No

[..few more examples..]

Goal: open the locked green door and go to the blue box
So far:  pick:green:key, unlock:green:door
Question: Should I  goto:blue:box?
Answer:
\end{lstlisting}

\subsubsection{SkillHack Prompts}
The NetHack environment is originally a text based game., So luckily we get the language description of our action and observations from the game engine.

\begin{lstlisting}[gobble=0, language=plantuml, breakautoindent=false, frame=trBL]
You are a NetHack Agent equipped with the following skills: 

ApplyFrostHorn:  Use a frost horn to freeze some lava.
Eat: Eat an apple.
Fight: Hit a monster.
NavigateLava: Reach the staircase past random lava patches.
PickUp: Pick up a random item.
PutOn: Put on an amulet or towel.
TakeOff: Take off clothes.
Unlock: Use a key to unlock a locked door.
Wear: Wear a robe.
Wield: Wield a sword.
ZapWandOfCold: Use a wand of cold to freeze lava.

<list common item names>

Battle: PickUp a randomly placed Sword, Wield the Sword and finally Fight and kill a Monster.

Goal: Battle the Monster
So far: I see a silver saber
Question: Should I TakeOff?
Answer: No

Goal: Battle the Monster
So far: picked up a silver saber
Question: Should I Wield?
Answer: Yes

Goal: Battle the Monster
So far: picked up a silver saber, weild a silver saber
Question: Should I ZapWandOfCold?
Answer: No

[..few more examples..]

Goal: Battle the Monster
So far: picked up sword, weild sword
Question: Should I Fight?
Answer:
\end{lstlisting}

\begin{lstlisting}[gobble=0, language=plantuml, breakautoindent=false, frame=trBL]
You are a NetHack Agent equipped with the following skills: 

[..same as above..]

Frozen Lava Cross: Either a Wand of Cold or a Frost Horn will spawn on the near side of a river of lava. PickUp the item and then create a bridge across the lava with either ZapWandOfCold or ApplyFrostHorn. Finally, NavigateLava across your newly made bridge to reach the staircase on the other side.

<list common item names>

Goal: Reach the staircase past the lava
So far: I see a wand
Question: Should I PickUp?
Answer: Yes

[..few more examples..]

Goal: Reach the staircase past the lava
So far:  picked up a wand, apply zap, the lava cools and solidifies
Question: Should I NavigateLava?
Answer: 
\end{lstlisting}

\subsubsection{Crafter Prompts}
For this environment, the prompts are designed similar to \cite{du2023guiding}.

\end{document}